\pdfoutput=1

\documentclass[11pt]{article}

\usepackage[final]{acl}

\definecolor{AccentBlue}{RGB}{0,122,204}
\definecolor{AccentGreen}{RGB}{70,160,73}

\newcommand{\prosodycommand}[2]{\newcommand{#1}{{\textcolor{AccentBlue}{\ensuremath{#2}}\xspace}}}
\prosodycommand{\Prosody}{\mathrm{P}_t}
\prosodycommand{\prosody}{\mathrm{p}_t}
\prosodycommand{\prosodyprime}{\mathrm{p}_t'}
\prosodycommand{\prosodyN}{\mathrm{p}_t^{(n)}}
\prosodycommand{\prosodyspace}{\mathcal{P}}

\newcommand{\wordcommand}[2]{\newcommand{#1}{{\textcolor{AccentGreen}{\ensuremath{#2}}\xspace}}}
\wordcommand{\Word}{\mathrm{W}}
\wordcommand{\word}{\mathrm{w}}
\wordcommand{\Wordt}{\mathrm{W}_t}
\wordcommand{\wordsN}{\boldsymbol{\mathrm{w}}^{(n)}}
\wordcommand{\Words}{\boldsymbol{\Word}}
\wordcommand{\words}{\boldsymbol{\word}}
\wordcommand{\wordsprime}{\boldsymbol{\word}'}
\newcommand{\alphabet}{\mathcal{W}}

\newcommand{\WordsContextBase}[3]{\textcolor{AccentGreen}{\ensuremath{\boldsymbol{#3_{\overset{#1,#2}{\leftrightarrow}}}}}\xspace}

\newcommand{\WordsContext}[2]{\WordsContextBase{#1}{#2}{\Words}}
\newcommand{\wordsContext}[2]{\WordsContextBase{#1}{#2}{\words}}

\newcommand{\defn}[1]{\textbf{#1}}
\newcommand{\R}{\mathbb{R}}

\newcommand{\mi}{{\ensuremath{\mathrm{MI}}}\xspace}
\newcommand{\entropy}{\ensuremath{\mathrm{H}}\xspace}
\newcommand{\xent}{\ensuremath{\entropy_{\vtheta}}\xspace}

\newcommand{\dataset}{\mathcal{D}}
\newcommand{\testset}{\dataset_{\mathrm{tst}}}
\newcommand{\trainset}{\dataset_{\mathrm{trn}}}

\newcommand{\vtheta}{\boldsymbol{\theta}}
\newcommand{\ptheta}{\ensuremath{p_{\vtheta}}\xspace}

\newcommand{\fzero}{\ensuremath{f_0}\xspace}

\newcommand{\langmodel}{\mathrm{LM}}

\usepackage{times}
\usepackage{latexsym}

\usepackage[T1]{fontenc}

\usepackage[utf8]{inputenc}

\usepackage{microtype}

\usepackage{inconsolata}

\usepackage{amsfonts}
\usepackage{amsmath}
\usepackage{mathtools}
\usepackage{xfrac}
\usepackage{amsthm}
\usepackage{xspace}

\usepackage{graphicx}

\usepackage{xspace}
\usepackage{amsfonts}
\usepackage{amsthm}
\usepackage{amsmath}
\usepackage{amssymb}
\usepackage{cleveref}
\usepackage{bbm}
\usepackage{caption}
\usepackage{subcaption}
\usepackage{xcolor}
\usepackage{thm-restate}
\usepackage{booktabs}
\usepackage{enumitem}
\usepackage{multirow}
\usepackage{hyperref} 
\usepackage{scalerel}

\usepackage[textsize=tiny]{todonotes}

\newtheorem{hypothesis}{Hypothesis}

\usepackage{cleveref}
\crefname{section}{\S}{\S\S}
\Crefname{section}{\S}{\S\S}
\crefname{table}{Tab.}{}
\crefname{figure}{Fig.}{Figs.}
\crefname{algorithm}{Alg.}{}
\crefname{equation}{Eq.}{Eqs.}
\crefname{appendix}{App.}{}
\crefname{hypothesis}{Hypothesis}{}
\crefname{thm}{Theorem}{}
\crefname{prop}{Proposition}{}
\crefname{defin}{Definition}{}
\crefname{reduction}{Reduction}{}
\crefname{cor}{Corollary}{}
\crefname{observation}{Observation}{}
\crefname{assumption}{Assumption}{}
\crefname{lemma}{Lemma}{Lemmas}
\crefformat{section}{\S#2#1#3}
\crefformat{footnote}{#2\footnotemark[#1]#3}

\title{The time scale of redundancy between prosody and linguistic context}

\author{Tamar I. Regev,$^1$ Chiebuka Ohams,$^1$ Shaylee Xie,$^2$ Lukas Wolf,$^3$ \\ 
\textbf{Evelina Fedorenko,$^1$ Alex Warstadt,$^4$ Ethan G. Wilcox,$^5$ Tiago Pimentel$^3$} \\
  $^1$MIT, $^2$Wellesley College, $^3$ETH Z\"urich, $^4$UCSD, $^5$Georgetown \\
  \texttt{\{tamarr,cohams,evelina9\}@mit.edu}, \texttt{sx102@wellesley.edu}, \\
  \texttt{\{lukas.wolf,tiago.pimentel\}@inf.ethz.ch}, \\
  \texttt{awarstadt@ucsd.edu}, \texttt{ethan.wilcox@georgetown.edu}}

\begin{document}
\maketitle
\begin{abstract}

In spoken communication, information is transmitted not only via words, but also through a rich array of non-verbal signals, including prosody---the non-segmental auditory features of speech. Do these different communication channels carry distinct information?
Prior work has shown that the information carried by prosodic features is substantially redundant with that carried by the surrounding words.
Here, we systematically examine the time scale of this relationship, studying how it varies with the length of past and future contexts. 
We find that a word's prosodic features require an extended past context (3-8 words across different features) to be reliably predicted.
Given that long-scale contextual information decays in memory, prosody may facilitate communication by adding information that is locally unique. 
We also find that a word’s prosodic features show some redundancy with future words, but only with a short scale of 1-2 words, consistent with reports of incremental short-term planning in language production. 
Thus, prosody may facilitate communication by helping listeners predict upcoming material.
In tandem, our results highlight potentially distinct roles that prosody plays in facilitating integration of words into past contexts and in helping predict upcoming words.

\vspace{0.5em}
\includegraphics[width=1.25em,height=1.25em]{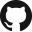}\hspace{1em}\parbox{\dimexpr\linewidth-2\fboxsep-2\fboxrule}{\url{https://github.com/Chief-Buka/contextual-redundancy}}

\end{abstract}

\section{Introduction}
\label{sec:intro}

Auditory features of speech such as pitch, loudness, and tempo---collectively termed prosody---play a crucial role in conveying meaning. Prosody influences sentence-level interpretation, encoding both linguistic and para-linguistic cues relevant to the communicative context \cite{cole2015prosody, wagner2010experimental, breen2010acoustic}. For example, prosody can signal phrase boundaries, emphasize key elements, transform statements into questions, and express sarcasm, excitement, or doubt. 
However, much of the information conveyed by prosody is redundant with the information encoded in the words themselves \cite{wolf2023quantifying}---that is, it is possible to predict a word's prosodic features from its \defn{linguistic context}, i.e., its surrounding words.\footnote{In both \citet{wolf2023quantifying} and our study, text is used as a proxy to measure information present in the words themselves, which we refer to as "linguistic information"(also termed ``segmental information'' in the phonology literature).} 

\begin{figure}[t]
    \centering
   \includegraphics[trim={0 .5cm 0 0cm},clip,width=\columnwidth]{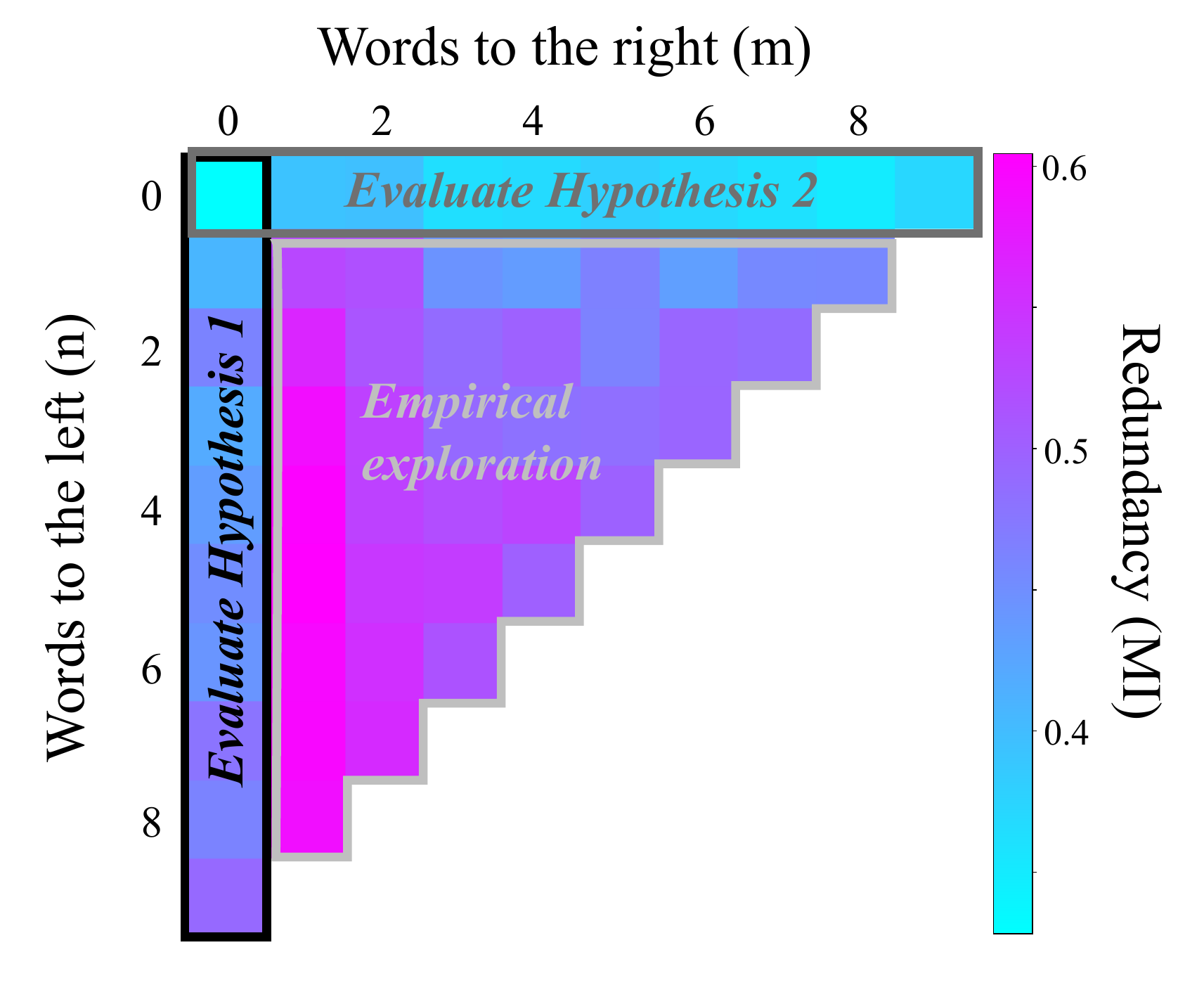}
   \caption{
   Redundancy (\mi) between prosody (\Prosody)  and linguistic context (\WordsContext{n}{m})
   as a function of the number of words contained in the past ($n$) and/or future ($m$) context. 
   The values are averaged across 6 prosodic features (see \cref{fig:results} for each of these features separately). 
   The first column is used to evaluate \cref{hyp:past}; the first row, \cref{hyp:future} (see section \cref{sec:intro}). Time scales are defined as the time after which the \mi does not increase. 
   }
   \label{fig:average}
   \vspace{-7pt}
\end{figure} 

Past work, however, only quantified the redundancy between prosody and the \emph{entire} linguistic context.
The question of the \emph{dynamics} of this redundancy is thus left open.
Here, we investigate the time scale of redundancy between prosodic features of a given word and its linguistic context by systematically varying the length of past or future context from 0 to 9 words in each direction (0 includes only the word whose prosody is being tested).

\emph{A priori}, what might one expect regarding the time scale of redundancy between prosodic features and \defn{past linguistic context}? 
During language comprehension, context is often critical to interpreting the current word and integrating it into the linguistic representations built thus far \citep{hale2001probabilistic,levy2008expectation}.
However, human memory is limited: contextual information decays over time, and connecting an incoming word to one far back is costly \cite{gibson1998linguistic, lewis2006computational, futrell2020lossy}. Given these memory limitations, prosody may be critically helpful in conveying information that is redundant with past context but is \emph{locally unique}, such that if a listener's representation of the past context is imperfect, they would benefit from the prosodic information contained by the word itself. We therefore hypothesize that prosody carries information that is recoverable from \emph{long-term past context}, but not from the short-term past (see \cref{hyp:past} below). 

\newcommand{\addcites}{{\color{red}(add cites)}\xspace}

With respect to \textbf{future linguistic context}, lexical and syntactic planning in language production is known to be relatively restricted \cite{brown2008little,bock1994language}: we may have a general idea of what we want to say next, but the particular words and constructions are selected incrementally, often with revisions. 
As a result, how we pronounce a current word and what prosodic features we associate with it should not be affected by long-term future linguistic context.
We therefore hypothesize that prosody carries information that is not recoverable from long-term future context (see \cref{hyp:future} below).

The above logic is laid out in terms of ``short'' vs.\ ``long'' contexts, but what constitutes a short or long time scale during language processing? Following recent work in neuroscience \cite{jain2018incorporating,regev2024neural,shain2024distributed} as well as work investigating intonational units in prosody itself \citep{inbar2020sequences}, we set the cutoff between ``short'' and ``long'' at \emph{roughly} 1-2 words, although see section \Cref{sec:discuss_timescale} for more discussion.

We thus formulate two main hypotheses:
\begin{hypothesis} \label{hyp:past}
    \defn{Long-scale past redundancy}. The redundancy between prosody and past linguistic context unfolds across a long time scale (longer than 1-2 words).
\end{hypothesis}

\begin{hypothesis} \label{hyp:future}
    \defn{Short-scale future redundancy}. The redundancy between prosody and future linguistic context unfolds across a short time scale (of 1-2 words).
\end{hypothesis}

To test these hypotheses, we quantify the redundancy between prosody and linguistic context as their mutual information, which we estimate using pre-trained language models, following \citet{wolf2023quantifying}.
Notably, we extend their approach to investigate the time scale of this redundancy by jointly varying context lengths parametrically, from 0 to 9 words for both past and future contexts.\footnote{In addition to examining the effects of the length of the past and future contexts separately, we also considered different combinations of past and future context lengths in order to identify the most predictive combination.}
We then analyze how redundancy changes across time for several commonly discussed prosodic features: pitch, loudness, duration, pause, and prominence. 

Our main results, averaged across all prosodic features considered are shown in \cref{fig:average} (see \cref{fig:results} for feature-specific results). In line with our hypotheses,
these results demonstrate that prosody is redundant with relatively long-term past linguistic context (up to 8 words), but is only redundant with short-term future context (up to 2 words). 
This finding advances our understanding of the dynamics of the redundancy between linguistic context and prosody, with potential implications for the role of prosody in natural spoken communication.

\section{Prosodic Features}
\label{sec:prosody}

Prosodic information is conveyed through multiple acoustic features of speech. Here, we examine the redundancy between linguistic context and: pitch, loudness, duration, pause and two versions of prominence (as detailed in section \cref{sec:dataset}). We chose these prosodic features as they have been commonly investigated in prior prosody research \citep[e.g., ][]{breen2010acoustic,gibson1998linguistic,cole2015prosody}. 

\paragraph{Pitch} 
Pitch is the perceptual dimension over which listeners order sounds on a scale from low to high. 
The acoustic correlate giving rise to this perception is the periodicity of sound signals;
pitch is thus often measured as the fundamental frequency (\fzero) of the sound.
Pitch is (arguably) the hallmark feature of prosody, 
having been extensively studied and characterized \citep{pierrehumbert1980phonology,silverman1992tobi,jun2006prosodic}. 
In stress-accent languages like English, pitch contours carry contextual information that can signal a wealth of information, including emphasizing specific words, signaling boundaries, speech act type, and the speaker's intent (e.g., interrogation, sarcasm, or affective state). Some typical pitch curves are the rise of pitch towards the last word of a question in yes/no questions in American English, the rise and then fall of pitch on a specific word to emphasize it, and a fall toward the end of phrases. Note that we do not study here the relationship between pitch and lexical identity, as in tonal-languages. However, see \citet{wilcox2025using}, who use similar methods to study typological pitch variation across languages.

\paragraph{Loudness} Loudness is the perceptual dimension over which listeners can order sounds on a scale from quiet to loud. The acoustic correlate giving rise to this perception is sound pressure, being measured as the intensity of acoustic energy.
The loudness of speech can be used to transmit information, such as to emphasize important words or convey emotion. The correlation between pitch and loudness is partly explained by vocal production constraints; producing speech with higher energy helps raise and better control the fundamental frequency. However, loudness variations may also convey independent information from pitch. 

\paragraph{Duration}
A word's duration is the difference between its offset (end) and onset (start) times.
The relationship between word duration and linguistic information has long been studied as a signature of efficiency in communication, such that more predictable words are reduced to a shorter duration \citep{jurafsky1998reduction, bell2009predictability, seyfarth2014word,coupe2019different,pimentel-etal-2021-surprisal}. Further, elongating a word is a common way to emphasize it, or signal prosodic boundary. Duration is thus also highly correlated with pitch and loudness in natural speech, but it can also be used independently to convey meaning, or to compress words of low information content.

\paragraph{Pause}
A pause following a word is the time difference between the word's offset (end) time and the next word's onset (start), being
another way to emphasize an important word in context, or to signal phrase boundaries \citep{hawkins1971syntactic}.
In contrast to phrase boundaries, within the phrase speech tends to be `connected' such that there are usually no pauses between words; i.e., most pauses are of zero seconds (see section   \cref{sec:dataset}).

\paragraph{Prominence}

Prosodic prominence is a term that describes how salient a linguistic entity---in our case, a single word---is perceived relative to the words surrounding it in an utterance \cite{terken2000perception}. Unlike the previously described prosodic features, prominence is a higher-level percept in the sense that it is not elicited by a single acoustic dimension. The perception of prominence is affected by multiple acoustic features \cite{cole2010signal}---elongating a word, increasing the speech energy or modulating the \fzero contour of the word can all make this word be perceived as more prominent in context. Although other acoustic features, like timbre, can also affect prominence, and factors like word frequency influence it as well \cite{cole2010signal}, a combination of duration, loudness and pitch has been proposed as an effective acoustic measure to quantify prosodic prominence \cite{talman2019predicting}, which we use here.

\vspace{5pt}
Notably, the prosodic features above present a high degree of correlation in natural speech \cite{ladd1996intonational}. 
For instance, as noted above, producing a higher pitch generally requires greater vocal effort, resulting in increased intensity; similarly, it may take longer to reach a higher pitch target, creating dependencies between pitch, loudness, and duration. 
Furthermore, prominence is strongly correlated to pitch, loudness and duration, as it is directly computed using a combination of these acoustic measures. 
We did not model these correlations explicitly; instead, we quantified the redundancy of each feature with linguistic context separately. 
This design choice does not affect our central research question, which concerns the time scale of these redundancies.
Furthermore, although these prosodic features are correlated, the differences observed across them in our results may indicate that they also carry distinct information.

\vspace{-3pt}
\section{Redundancy between Prosody and Linguistic Context}

This paper concerns the time scale of the redundancy between prosody and linguistic context, where `linguistic context' here refers to the segmental information of an utterance, represented in our experiments as text.
%
In this section, we first explain how this redundancy can be formalized as a mutual information, 
following \citet{wolf2023quantifying}.
We then expand on this framework by discussing how context-length manipulations allow us to investigate the time scale of this redundancy.

\subsection{Redundancy as Mutual Information}

Let $\Prosody$ be a prosody-valued random variable, which takes values $\prosody \in \R$.
Further, let $\Words$ be a words-valued random variable, which takes values $\words\! \in\! \alphabet^*\!\!$, where $\alphabet$ is a language's lexicon.
We follow \citet{wolf2023quantifying} in formalizing the redundancy between prosody and linguistic context as the mutual information: $\mi(\Prosody; \Words)$.
Under a few technical assumptions (e.g., the good mixed-pair assumption, see \citeauthor{wolf2023quantifying} for details), we can write this as:
\begin{align}
    \mi(\Prosody; \Words) &= \entropy(\Prosody) - \entropy(\Prosody \mid \Words) 
\end{align}
In this equation, \defn{unconditional entropy} $\entropy(\Prosody)$ serves as a baseline, representing how much uncertainty there is about $\Prosody$, in general.
In turn, \defn{conditional entropy} $\entropy(\Prosody \mid \Words)$ represents how much uncertainty remains about $\Prosody$ once we know $\Words$.
Their difference then represents how much information $\Words$ contains about $\Prosody$ (and \textit{vice versa}).

We are now left with the problem of estimating these entropies.
While these values are unknown, we only require two things to estimate them: a corpus of prosodic values coupled to linguistic contexts, sampled from the ground-truth distribution,
\begin{align}
    \testset = \{\prosodyprime, \wordsprime\}_{n=1}^{N},\,\, \prosodyprime, \wordsprime \sim p(\prosody, \words) \nonumber
\end{align}
and models $\ptheta$ of distributions $p(\prosody)$ and $p(\prosody\mid\words)$.
We can then use a cross-entropy upper-bound \cite{pimentel-etal-2019-meaning} to estimate these entropies:
\begin{align} \label{eq:xentropy}
    \entropy(\Prosody\mid\Words) &\leq \xent(\Prosody \mid \Words) \\
    &\approx \frac{1}{|\testset|} \sum_{\prosodyprime, \wordsprime \in \testset}\!\! \log \frac{1}{\ptheta(\prosodyprime \mid \wordsprime)} \nonumber
   \end{align}
where $\ptheta(\prosody \mid \words)$ is replaced with $\ptheta(\prosody)$ when estimating $\entropy(\Prosody)$.
We describe our dataset $\testset$ and how to estimate $\ptheta$ in section \Cref{sec:methods}.

\subsection{Manipulating Context Length}

\begin{figure}[t]
   \includegraphics[trim={0 0 0 1.3cm},clip,width=\columnwidth]{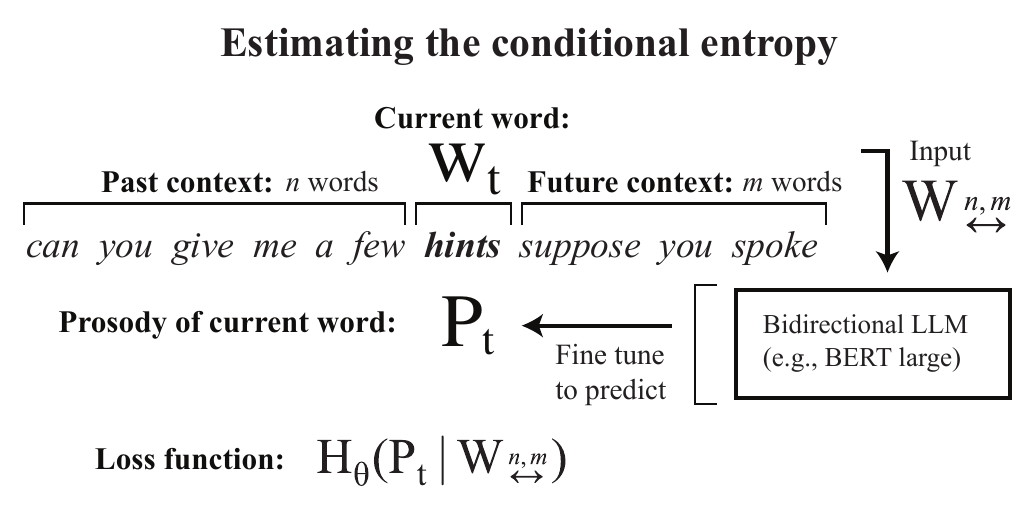}
   \vspace{-20pt}
   \caption{
   Estimation procedure for $\entropy(\Prosody \mid \WordsContext{n}{m})$.
   A span of words \WordsContext{n}{m} which includes word $\Wordt$ is used as input to a model which predicts that word's prosody $\Prosody$.
   The loss function that the model minimizes estimates the conditional entropy.}
   \label{fig:method}
   \vspace{-8pt}
\end{figure}

To analyze the time scale of the redundancy between prosody and linguistic context, we will estimate \mi(\Prosody; \Words) while systematically manipulating the amount of context, i.e. the number of words, in \Words. We thus quantify `time' in units of words, as opposed to seconds, acknowledging the discrepancy between these concepts due to varying duration of words and speaking rates.
To this end, we define \WordsContext{n}{m} as the linguistic context comprising $n$ words before and $m$ words after word $\Wordt$, including the word itself:
\begin{align}
    \wordsContext{n}{m} = \langle \word_{t-n}, \cdots, \word_{t}, \cdots \word_{t+m} \rangle
\end{align}
Thus, for instance, \WordsContext{0}{0} corresponds to the word $\Wordt$ by itself, and \WordsContext{5}{3} corresponds to the word $\Wordt$ with 5 words in its past context and 3 words in its future context (see \cref{fig:method}).

Given this definition, we can then explore the time scale we are interested in by estimating the mutual information $\mi(\Prosody, \WordsContext{n}{m})$ while varying $n$ and $m$. 
This amounts to estimating $\entropy(\Prosody)$ and $\entropy(\Prosody \mid \WordsContext{n}{m})$.
The unconditional entropy does not depend on context; thus, to estimate it, we only need to compute a distribution over the domain of each prosodic feature.
On the other hand, estimating the conditional entropy requires computing a family of conditional distributions for each prosodic feature, with one distribution for each $n, m$ combination.
In other words, we need a model $\ptheta(\prosody \mid \wordsContext{n}{m})$ that works for any $n,m$ pair.
We elaborate on this model in section \cref{sec:estimating}.

Importantly, the $\mi(\Prosody; \WordsContext{n}{m})$ is a monotonically increasing function of both $m$ and $n$; larger contexts must contain at least as much (but maybe more) information about prosody than smaller contexts.\footnote{We note that---while the MI is monotonically increasing in theory---it is not necessarily the case that this underlying monotonicity will be reflected by our estimation methods.}
However, at some value of $m$ and some value of $n$, the MI might reach a plateau.
We consider the \defn{time scale} of the redundancy between prosody and linguistic context to be the value from which increasing context length does not significantly increase the MI.

\section{Methods} \label{sec:methods}


\subsection{Dataset}
\label{sec:dataset}

Our data-extraction process uses \citeposs{wolf2023quantifying} proposed and publicly available pipeline (see their paper for more details).

\paragraph{Speech Data} We use the LibriTTS spoken language corpus 
\cite{zen2019libritts},\footnote{This dataset is derived from LibriSpeech audiobooks corpus \cite{panayotov2015librispeech}, which is itself derived from LibriVox \cite{kearns2014librivox}.} 
which contains public domain audiobook materials (audio and text) recorded by volunteer narrators.
This dataset contains 585 hours of English speech data at a 24kHz sampling rate, and includes recordings from 2,456 speakers reading aloud books which are paired with the corresponding transcripts. We filtered out texts from LibriTTS that contained less than three words (such as book and chapter titles) and we eliminated punctuation marks, since these can be very informative regarding prosody, and are not explicitly present in spoken communication.

\paragraph{Prosody Feature Extraction} 
For extracting prosodic features we began by aligning the audio and text using the Montreal Forced Aligner \citep[MFA;][]{mcauliffe2017montreal}.
With the alignment in place, both the duration and pause of each word can be easily computed from the words' offset and onset times. Notably, $89.4\%$ of the words in our dataset have a pause of 0 seconds. 
Duration was normalized by the number of syllables (from CELEX; \citealp{baayen1996celex}) to obtain duration per syllable, reducing variability across word identities.
 
Pitch and loudness 
were extracted using the algorithms from \citet{suni2017hierarchical}.
These yield pitch---measured as fundamental frequency, \fzero---curves that we $z$-scored per speaker to normalize for inter-speaker pitch range and avoid bimodal pitch distributions due to sex.
For each word, we analyzed pitch in a window of up to 250$ms$ (or the word's full duration, if shorter) centered on the primary (stressed) syllable, as identified using CELEX. This choice reflects the fact that English pitch accents typically align with stressed syllables \cite{pierrehumbert1980phonology} and reduces phonological variability due to word length. However, this approach focuses on lexical stress and may underrepresent boundary tones. We then averaged the pitch values within the selected window to yield a single mean pitch value per word. 
Loudness, measured as energy, was similarly extracted, with energy curves being averaged per word.

Prominence was computed using a composite acoustic measure from \citet{suni2017hierarchical}, which integrates time-frequency variation across duration, energy, and \fzero. This measure has been validated in prior work as capturing prosodic prominence \cite{suni2017hierarchical, talman2019predicting, terken2000perception}. 
In addition to the absolute prominence values, we also computed relative prominence by subtracting the mean prominence of the preceding three words from the current word’s value, emphasizing local contrast. 
We used the word-level prominence scores released with this dataset \cite{talman2019predicting}.\looseness=-1

\paragraph{Splitting the Data into Train, Validation and Test} The dataset was divided into train, validation and test sets, following standard practice in machine learning. We used splits of the dataset provided by \citet{talman2019predicting}. For training, we used a data split (termed train-360) containing 904 speakers, 11,262 sentences and 2,076,289 words. For validation, we used a data split (termed dev) containing 40 speakers, 5,726 sentences and 99,200 words. For testing, we used two data splits (both non-overlapping with the Train and Validation sets): one (termed train-100) containing 247 speakers, 33,041 sentences and 570,592 words, and the other (termed test) containing 39 speakers, 4,821 sentences and 90,063 words. 
We used the train-100 split as the main test set, except when analyzing prominence; for those analyses, we combined the two test sets in order to obtain more stable results.

\subsection{Estimating the Cross-Entropies}
\label{sec:estimating}

We now explain how we model the probability distributions $\ptheta(\prosody)$ and $\ptheta(\prosody \mid \wordsContext{n}{m})$, which serve to estimate the unconditional and conditional cross-entropies, respectively. 

\newcommand{\sigmatrain}{\widehat{\sigma}}

\paragraph{Modeling $\ptheta(\prosody)$}
To model this unconditional distribution over prosodic values, we simply follow \citet{wolf2023quantifying} in using a Gaussian kernel density estimator (KDE).
Given a training set $\trainset$, sampled from $p(\prosody, \words)$, this model is defined as:
\newcommand{\normaldist}{\mathcal{N}}
\begin{align}
    &\ptheta(\prosody) =  \frac{1}{|\trainset|} 
    %
    \!\sum_{\prosodyprime \in \trainset}\!\!
    \normaldist(\prosody; \mu\!=\!\prosodyprime, \sigma\!=\!\sigmatrain)
\end{align}
where $\normaldist$ are Gaussian distributions, each centered at a prosodic value $\mu\!=\!\prosodyprime$ and all having the same variance $\sigma\!=\!\sigmatrain$.
We choose $\sigmatrain$ that achieves the highest likelihood on our validation set.


\newcommand{\paramdist}{\mathcal{Z}}
\newcommand{\paramdistparams}{\boldsymbol{\phi}}
\newcommand{\paramdistparamsfit}{\widehat{\boldsymbol{\phi}}}

\paragraph{Modeling $\ptheta(\prosody \mid \wordsContext{n}{m})$}
To model this conditional distribution, we again follow \citeauthor{wolf2023quantifying} in
finetuning a language model ($\langmodel$), with an added linear layer on top, to predict the parameters of a conditional distribution over $\prosody$. Using language models to represent text offers greater expressivity than traditional regression-based approaches.
Unlike \citeauthor{wolf2023quantifying}, however, we limit our model's input to include only $\wordsContext{n}{m}$  instead of the entire $\words$ (\citeauthor{wolf2023quantifying} used the entire available context in the segments from the LibriTTS corpus, which varied in length between 4 and 60 words).

We assume the conditional distribution over prosody follows a parametric distribution $\paramdist$, and use a $\langmodel$ to predict this distribution's parameters $\paramdistparamsfit$:\footnote{We evaluate models with Gaussian, Gamma and Laplace distributions, choosing the distribution that leads to the lowest cross-entropy on a validation set. Parameters $\paramdistparams$ are, e.g., the mean and standard deviation for a Gaussian.
}\looseness=-1
%
\begin{align}
    &\paramdistparamsfit = \langmodel(\wordsContext{n}{m}) \\
    &\ptheta(\prosody \mid \wordsContext{n}{m}) = 
    \paramdist(\prosody; \paramdistparams=\paramdistparamsfit)
\end{align}
We finetune this model by minimizing its cross-entropy on a training set $\trainset$; which amounts to minimizing the right-hand side of \cref{eq:xentropy}.
As the cross-entropy is an upper bound on the entropy, the lower its value (and consequently the better our model) the tighter the estimate we get for the entropy.
Due to compute time considerations, we estimate all $n,m$ combinations using a single finetuned $\langmodel$: 
During training, we sampled
inputs of varying lengths, spanning 1 to 10 words. These inputs were obtained by first sampling an item from the dataset, and then randomly cutting it into the desired length (between 1-10 words).
For each sample, the model then predicts the prosody of each of the words in this span in parallel; in a 7-word span, thus, the first word's prosody is predicted as $\ptheta(\prosody \mid \wordsContext{0}{6})$ and the 5th word's prosody is predicted as $\ptheta(\prosody \mid \wordsContext{4}{2})$. Importantly, we made sure that the models saw an equal number fo samples for each $n,m$ combination. 
For each prosodic feature, we tested several models, namely BERT, BERT-large and RoBERTa-large, and selected the model that gave the best results.
In most cases, this was BERT-large, except for pauses and pitch where it was BERT.
An early stopping criterion was applied, such that if the loss did not decrease for 3 epochs the model stopped training.

\begin{figure*}
   \centering
   \includegraphics[width=\textwidth]{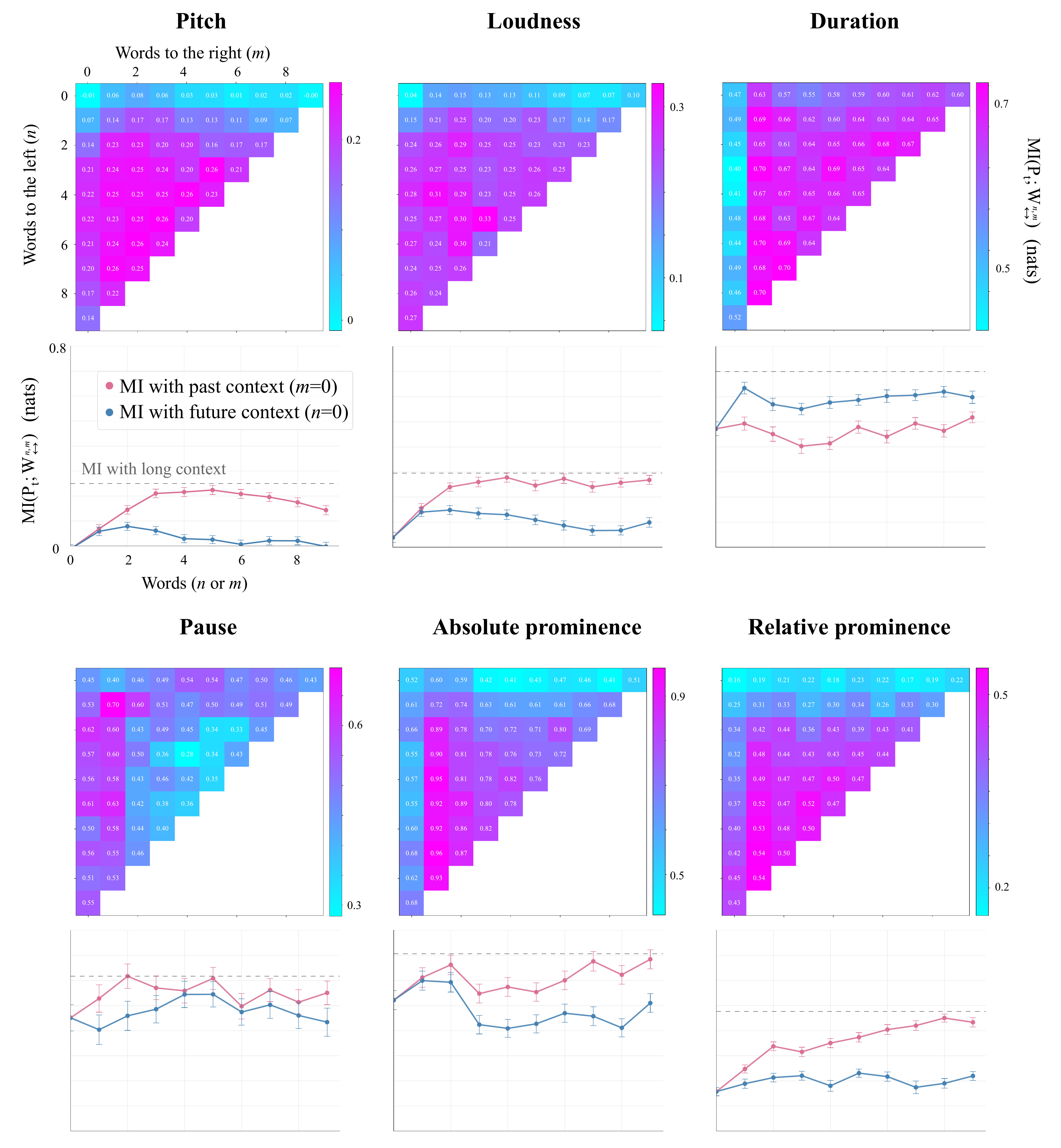}
   \caption{For each of the 6 tested prosodic features, two plots are presented. The upper plots are similar to \cref{fig:average}, and display the redundancy, quantified as mutual information between a prosodic feature at a given word \prosody{} and the linguistic context \wordsContext{n}{m}, which includes the word itself and a varying length context, $n$ words before, and $m$ words after it. 
   The lower plots display either: (a) in red, just the first column from the upper plots (corresponding to past linguistic context, which includes the word itself and a gradually increasing number of previous words, red curve) and (b), in blue, the first row from the upper plots (corresponding to future linguistic context, which includes the word itself and a gradually increasing number of future words, blue curve). Details displayed only for Pitch or Duration (labels, numbers, legend) are the same for all features. The MI values correspond to the mean across all test data. Error bars correspond to standard errors of the mean. Dashed horizontal gray lines corresponds to the MI with the longest available context, taken from \citet{wolf2023quantifying} (between 6 and 60 words into the past or future). See section \cref{sec:histograms} for the distributions of these features in our dataset.
     }
   \label{fig:results}
   \vspace{-3.0pt}
 \end{figure*}

\section{Results}

We present our main results in \cref{fig:average,fig:results}.
Namely,
\cref{fig:average} displays the average mutual information (MI) across the 6 prosodic features tested here (pitch, loudness, duration, pause, absolute prominence, relative prominence), while \cref{fig:results} displays those results for each feature separately. 
See section \cref{sec:unconditional} for the values of the unconditional entropies comprising those MIs.

The overall trend is consistent with the hypotheses outlines above. 
In line with the long-scale past redundancy hypothesis (\cref{hyp:past}),
the mutual information between prosody (averaging across all prosodic features used here) and past context (first column in \cref{fig:average}) increases as a function of context length until up to about 5-8 words, at which point it plateaus.
In line with the short-scale future redundancy hypothesis (\cref{hyp:future}), the \mi between prosody (averaging across all features) and future context (first row in \cref{fig:average}) increases only until up to one or two words and then plateaus.
Furthermore, at context lengths above one word, the MI with the past context tends to be higher than the MI with the future context, with this trend becoming more pronounced at longer contexts.

We also empirically explore other combination of past and future contexts (i.e., $n>0$ and $m>0$; the $n$-th column and $m$-th row in \cref{fig:average}).
This analysis revealed that prosodic features at the current word are best predicted by a combination of 5-8 words in the past and about one word in the future.
Notably, this $n,m$ combination led to higher mutual information than even other combinations with larger context (i.e., with $n',m'$, $n'\!\geq\! n$, and $m'\!\geq\! m$).
While this is theoretically impossible (adding more context can never decrease mutual information), this is likely due to our models' training procedure not being able to ignore unhelpful contributions of long-scale contexts.

Interestingly, individual prosodic features show somewhat distinct patterns (see \cref{fig:results}). 
The long-scale past hypothesis (\cref{hyp:past}) is supported for most prosodic features individually, but not for duration and pauses. 
For pitch, loudness, and prominence, the MI with past context (red curves in the lower plot for each feature in \cref{fig:results}) increases until up to at least $3$ words (although the curve for absolute prominence is a little noisy).
For duration, however, the MI with past context never rises significantly above the $0,0$ point, indicating the past context does not add information beyond the word identity itself; and for pauses, the past MI saturates after two words.

The short-scale future hypothesis (\cref{hyp:future}) is also supported for most individual prosodic features, except for duration and pause. 
For pitch, loudness, and prominence, the MI with future context is shorter than with past context, saturating between 1 and 3 words (blue curves in the lower plot in each feature in \cref{fig:results}).
For duration, the MI with future context saturates after 1 word, which is longer than with past context (the past curve shows no increasing trend).
For pauses, the future MI saturates after about 4 words, which is longer than with past context---although these MI estimates are noisy for this feature, likely due to data sparsity.
Notably however, the MI with past context is higher than with future context for all prosodic features except for duration.
We thus conclude that the results from these individual prosodic features generally support the short-scale future hypothesis.


\section{Discussion}

This work aimed to characterize the time scale of the redundancy between prosodic and linguistic information.
We built on the foundation of past work, which established that linguistic information is partially redundant with prosodic features \cite{wolf2023quantifying}, to systematically manipulate the length of past and future context, 
from 1 to 10 words. 
Our extension of this pipeline is publicly available for researchers wishing to extend our analysis, potentially exploring the time scale of redundancy between other communication channels of interest.

Based on what is known about human memory limitations and linguistic planning, we hypothesized that redundancy between linguistic information and prosodic features extends over a relatively long time scale into the past, and a relatively short time scale into the future (see section \Cref{sec:intro}). We find support for these  hypotheses: 
For most prosodic features we tested (apart from duration and pause, see section \cref{sec:discuss_pause_dur}), redundancy with past linguistic context unfolds across a long time scale (of 3-8 words), whereas redundancy with future linguistic context is shorter-scale (of 1-2 words). 
We next discuss the implications of these results with respect to the previous literature.

\subsection{The Time Scales of Linguistic and Prosodic Information}\label{sec:discuss_timescale}

Sentence comprehension is constrained by cognitive demands such as attention and working memory, leading to lossy representations of past linguistic context \citep{gibson1998linguistic,vasishth2010short,futrell2020lossy,kuribayashi2022context}.
It is, however, challenging to estimate the precise number of words maintained in working memory during language processing, as this likely depends on many factors. 
Neural evidence suggests that language-selective neural populations\footnote{Language-selective neural populations are clusters of neurons in the human brain that respond selectively to linguistic experimental manipulations, but not to manipulations in other perceptual or cognitive domains such as math and music. These populations are thought to constitute a brain network specialized for language processing \cite{fedorenko2024language}.} integrate linguistic information across time scales varying between 1-6 words \citep{jain2018incorporating,shain2024distributed,regev2024neural}.
Moreover, work in prosody has put forward the notion of "intonation units", which follow a rhythmic structure of about 1 Hz \citep{inbar2020sequences}.
Given the average speech rate of ~200 words per minute \citep{yuan2006towards}, each intonation unit contains about 3-4 words, which aligns with the relatively short context of linguistic processing.\footnote{
To be precise about the generality of the claims above, we note that:
the  claims in \citep{gibson1998linguistic,vasishth2010short,futrell2020lossy,kuribayashi2022context} are intended as universal and were shown for English, German, and Japanese;
the experiments in \citep{jain2018incorporating,shain2024distributed,regev2024neural} are in English;
the experiments in \citep{inbar2020sequences} cover 6 typologically diverse languages;
the experiments in \citep{yuan2006towards} are in English and Chinese.
}\looseness=-1

Our findings are generally in accord with these previous studies, and suggest that redundancy with past context spans around 3-8 words---a scale comparable to both the time scale of linguistic integration and prosodic segmentation. 

\subsection{Prosody as an Audience-Design Tool in Communication}

A longstanding debate in linguistics concerns the extent to which language production is shaped by audience design considerations, with speakers tailoring their utterances to facilitate comprehension. 
Some evidence suggests that syntactic choices are not strongly adapted for listener needs but rather reflect the speaker's own constraints \cite{ferreira2008ambiguity,morgan2022still}. This apparent lack of audience design in syntax may stem from the rigid structural constraints imposed by the rules of the language (at least in English, which is used in much past work). 
In contrast, prosodic choices may offer greater flexibility for accommodating audience-centered considerations, allowing speakers to dynamically modulate pitch, loudness, and rhythm in real time.
This flexibility may afford prosody a larger role in audience design \cite{clark_poliak_regev_haskins_gibson_robertson_2025}, serving as a communicative channel that facilitates comprehension.
For example, prior work shows a trade-off between a word's duration and its information content 
\citep{jurafsky1998reduction,bell2009predictability,coupe2019different,pimentel-etal-2021-surprisal}, which is typically interpreted as arising to facilitate comprehension by more evenly distributing information across time \citep[known as the uniform information density hypothesis;][]{fenk1980konstanz,genzel-charniak-2002-entropy,levy2007speakers,meister-etal-2021-revisiting}.
Our findings may be taken as further evidence for the audience-design view of prosody: 
the fact that its redundancy with past context extends over long time scales suggests that prosody may help clarify the relationship between the current word and the broader preceding context, thus facilitating the extraction of the intended meaning. 
Given that prosody appears to carry locally unique information,
prosodic features may aid comprehension when representations of longer-scale past are lossy.

\subsection{The Relationship Between Prosody and Future Words}

Our findings indicate that the redundancy between prosody and future words is overall lower compared to past words. 
However, prosody still exhibits a strong relationship with the upcoming word or two. 
This short-range relationship may arise from motor constraints on prosodic articulation, from local linguistic dependencies such as fixed expressions, or from short-scale prosodic planning---consistent with theories of language production that emphasize incremental, short-term planning \cite{brown2008little,bock1994language}. 
In addition, prosody may help listeners form expectations about upcoming words through cues such as duration, pauses, or pitch modulations. 
This interpretation aligns with recent findings that humans are better at predicting upcoming words given spoken, compared to written, context \cite{botch2025sensory}.
Relatedly, prior research has shown that a word's duration correlates with its predictability given future context \citep{bell2009predictability}.
Along similar lines, reading times
also seem to correlate with features of upcoming words \citep[such as frequency, predictability and entropy;][]{roark-etal-2009-deriving,angele2015successor,van2017approximations}. 
Future work should investigate these potential mechanisms to clarify the role of prosody in forward-looking processing.

\subsection{Duration and Pause Show Weak Past, Strong Future Redundancy}\label{sec:discuss_pause_dur}

Compared to the other prosodic features, duration and pause stand out in their relatively short time scale of redundancy with both past and future words, as well as in their relatively high MI with future words. 
This pattern may suggest that lengthening a word and pausing after it primarily serves to prepare for the upcoming word, facilitating its processing by slowing the rhythm of speech.
Alternatively, this pattern may stem from sentence boundaries: As discussed above, pauses are most common after sentence-final words, which are also often elongated \cite{seifart2021extent, paschen2022final}. The high predictability of duration and pause from future context may therefore partly reflect our models' ability to detect a sentence-final vs. sentence-medial word, an important structural cue which could, in turn, help predict the values of these prosodic features. 
While we removed punctuation from the input, therefore preventing sentence-boundary information from being conveyed explicitly through punctuation marks, we did not lowercase all the data and therefore sentence-initial words are clearly detectable by their capitalization. Furthermore, sentence boundary cues could be inferred from the lexical content itself, since certain words tend to occur more frequently at the beginning or end of sentences.

\section{Conclusion}

Our findings reveal a fundamental asymmetry in the time scale of redundancy between prosody and linguistic information: while prosody exhibits redundancy with both past and future words, this relationship extends across a longer span for past words (3–8 words) than for future words (1–2 words). 
This suggests that prosody’s relationship with future words primarily reflects short-term effects such as next-word prediction, local word dependencies, or other production factors---future work should try to distinguish those explanations. 
In contrast, its relationship with past words operates over a broader scale, potentially serving to reinforce or highlight information that may be cognitively demanding for listeners to process in real-time communication. 
These results provide new insights into the role of prosody in spoken language.

\section*{Limitations}

Our study has several limitations that should be considered when interpreting the results.

\paragraph{Data-related Limitations.} The first set of limitations relates to the dataset used. 
Our dataset consists of audiobooks, which do not necessarily reflect natural prosody in real time communication, potentially affecting the generalizability of our findings. 
Redundancy may be higher in audiobooks than in spontaneous speech, because the text is written with the assumption that it must convey all necessary information without relying on prosody. 
We address this concern to some extent by removing punctuation marks, which serve as a substitute for prosody in written text. 
Another dataset-related limitation is the sample size. 
Furthermore, the syntactic structures used in audiobooks may differ from those in spontaneous speech. Because pause and duration often signal syntactic boundaries, they may be particularly sensitive to such structural differences. This may help explain why pause and duration did not follow our hypotheses.
Larger datasets may be required for more stable estimations, especially given that we compute 55 different values (for \wordsContext{n}{m}, $n$ and $m$ from 0 to 9).
Furthermore, our dataset is limited to the English language. Therefore, our findings are English-specific and may not generalize across languages, given known cross-linguistic variation in prosody. Future work should generalize this work to other typologically distinct languages \citep[see ][for work using similar methods to study diversity in prosodic typology]{wilcox2025using}.

\paragraph{Estimation-related Limitations.} 
The second set of limitations has to do with the estimation procedure. 
The mutual information we compute tries to approximate the true value, and is constrained by the quality of our models $\ptheta(\prosody \mid \wordsContext{n}{m})$. 
One of our modeling assumptions is the functional form of the conditional distribution of prosody given a context (namely, Gaussian, Gamma or Laplace distributions depending on the prosodic feature).
However, this parametric assumption may limit the model's performance and future work should explore alternative conditional distributions which may improve results (as done by, e.g., \citealp{wilcox2025using}).
This assumption is particularly violated for features that manifest different distributions; pause, for instance, takes a 0 value in $89.4\%$ of the data and may therefore be better modeled by a zero-inflated distribution.
Indeed, our results for pause seem particularly noisy. 
Additionally, to estimate \wordsContext{n}{m}, we provided the models with short segments of 1 to 10 words. 
However, the language models used here (e.g., BERT) were not pretrained on such short segments, but rather on longer spans of text; this might have impacted their efficiency in extracting the information from short segments. 
While finetuning likely mitigates this issue, it remains a potential limitation. 
Furthermore, we train a single model for all combinations of $n,m$, which does not guarantee that the model is optimal for each combination separately. 
Finally, we observed cases where the mutual information decayed for longer contexts, which contradicts expectations from information theory, as additional context can never reduce information.
This phenomenon likely stems from issues in training the models, which could be biased toward under-utilizing the available context for longer spans. 
Future work should address these limitations to refine our understanding of redundancy between prosody and linguistic information.

\section*{Acknowledgments}
The authors would like to acknowledge Cui Ding and Giovanni Acampa for technical assistance. TIR was supported by the Poitras Center for Psychiatric Disorders Research, McGovern Institute for Brain Research, MIT. 

\section*{Ethics Statement}
We foresee no potential ethical concerns or risks associated with this study, although we acknowledge the inherent risks in using any AI system. 

\bibliography{custom}

\newpage
\appendix


\section{Unconditional Entropies}
\label{sec:unconditional}

\begin{table}[h]
    \centering
    \resizebox{\linewidth}{!}{%
    \begin{tabular}{lc}
         \toprule
         \textbf{Prosodic Feature} & \textbf{Unconditional Entropy} \\
         \midrule
         Absolute Prominence & \phantom{-}0.536 \\
         Relative Prominence & \phantom{-}1.355 \\
         Energy & \phantom{-}0.815 \\
         Duration & -0.920 \\
         Pause & -5.193 \\
         $f_0$ & \phantom{-}3.469 \\
         \bottomrule
    \end{tabular}%
    }
    \caption{Unconditional entropies of each prosodic feature.}
    \label{tab:unconditional_entropies}
\end{table}


\section{Prosodic Features' Histograms}
\label{sec:histograms}

\begin{figure}[h]
   \includegraphics[width=\columnwidth]{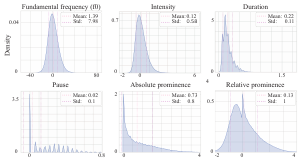}
   \caption{
   Histogram of prosodic features. All $y$-axes correspond to probability densities.
    (top-left) Pitch (fundamental frequency), units are arbitrary (z-scored);
    (top-center) Loudness, units are arbitrary (z-scored);
    (top-right) Duration, units are seconds per syllable;
    (bottom-left) Pause, units are seconds;
    (bottom-center) Absolute prominence, units are arbitrary;
    (bottom-right) Relative prominence, units are arbitrary.}
   \label{fig:histograms}
\end{figure}

\end{document}